\documentclass[letterpaper,10pt,journal,twoside]{IEEEtran}
\IEEEoverridecommandlockouts                              % 
\usepackage{times}  % DO NOT CHANGE THIS
\usepackage{helvet}  % DO NOT CHANGE THIS
\usepackage{courier}  % DO NOT CHANGE THIS
\usepackage[hyphens]{url}  % DO NOT CHANGE THIS
\usepackage{graphicx} % DO NOT CHANGE THIS
\usepackage{caption} % DO NOT CHANGE THIS AND DO NOT ADD ANY OPTIONS TO IT
\usepackage{algorithm}
\usepackage{algorithmic}
\usepackage{array}
\usepackage{subcaption}
\usepackage{multirow}
\usepackage{amssymb}
\usepackage{amsmath}
\usepackage{newfloat}
\usepackage{listings}
\usepackage{makecell}
\usepackage{placeins}
\newcommand{\dataset}{PotentialRiskQA}
\newcommand{\method}{PR-Reasoner}

\title{\LARGE \bf Predicting Risks Before They Happen: Vision-Language Reasoning for Autonomous Driving}

 \author{Jiaxin~Liu\textsuperscript{\rm 1*}, Xiangyu~Yan\textsuperscript{\rm 2*}, Liang~Peng\textsuperscript{\rm 1}, Lei~Yang\textsuperscript{\rm 1}, Yuechen~Luo\textsuperscript{\rm 1}, Lingjun~Zhang\textsuperscript{\rm 1}, Hangjun Ye\textsuperscript{\rm 3},\\ Guang Chen\textsuperscript{\rm 3}, Mu Li\textsuperscript{\rm 3}, Lei Zhang\textsuperscript{\rm 3}, Ziqi Zhan\textsuperscript{\rm 3}, Haiyang Sun\textsuperscript{\rm 3}, Sai Guo\textsuperscript{\rm 3}, Hong Wang\textsuperscript{\rm 1}, and Jun Li\textsuperscript{\rm 1} 
}

\begin{document}

\maketitle
\thispagestyle{empty}
\pagestyle{empty}

\begin{abstract}
% Waiting until a driving hazard becomes directly visible can leave an autonomous vehicle little time to respond. 
Observable precursors, such as another vehicle's unexpected braking, can provide evidence of a hidden hazard and an opportunity for earlier risk mitigation. However, existing driving datasets and language reasoning formulations provide limited supervision for connecting these precursors to potential risks and appropriate responses. We introduce PotentialRiskQA, an ego-view driving dataset that marks precursor and risk appearance times and provides structured annotations linking scene evidence, potential risks, and mitigation actions. Building on this dataset, PR-Reasoner trains vision-language models to generate a precursor-centered reasoning chain through scene-level supervision and auxiliary tasks for scene understanding, precursor identification, risk inference, driving suggestions, and meta-actions. Experiments across multiple backbones show improved scene understanding, risk cognition, and action quality, with fewer unsupported risk claims on most models. Ablations show that supervising intermediate reasoning steps further reduces unsupported risk claims and improves action quality beyond scene-level training alone. On NAVSIM, the trained representations improve open-loop planning performance, demonstrating benefits beyond language-based evaluation.

\end{abstract}

% !TeX root = ../RAL_main.tex

\section{Introduction}
\label{sec:intro}
\begin{figure*}[t]
    \centering
    \includegraphics[trim=0 113 0 8, clip, width=0.95\textwidth]{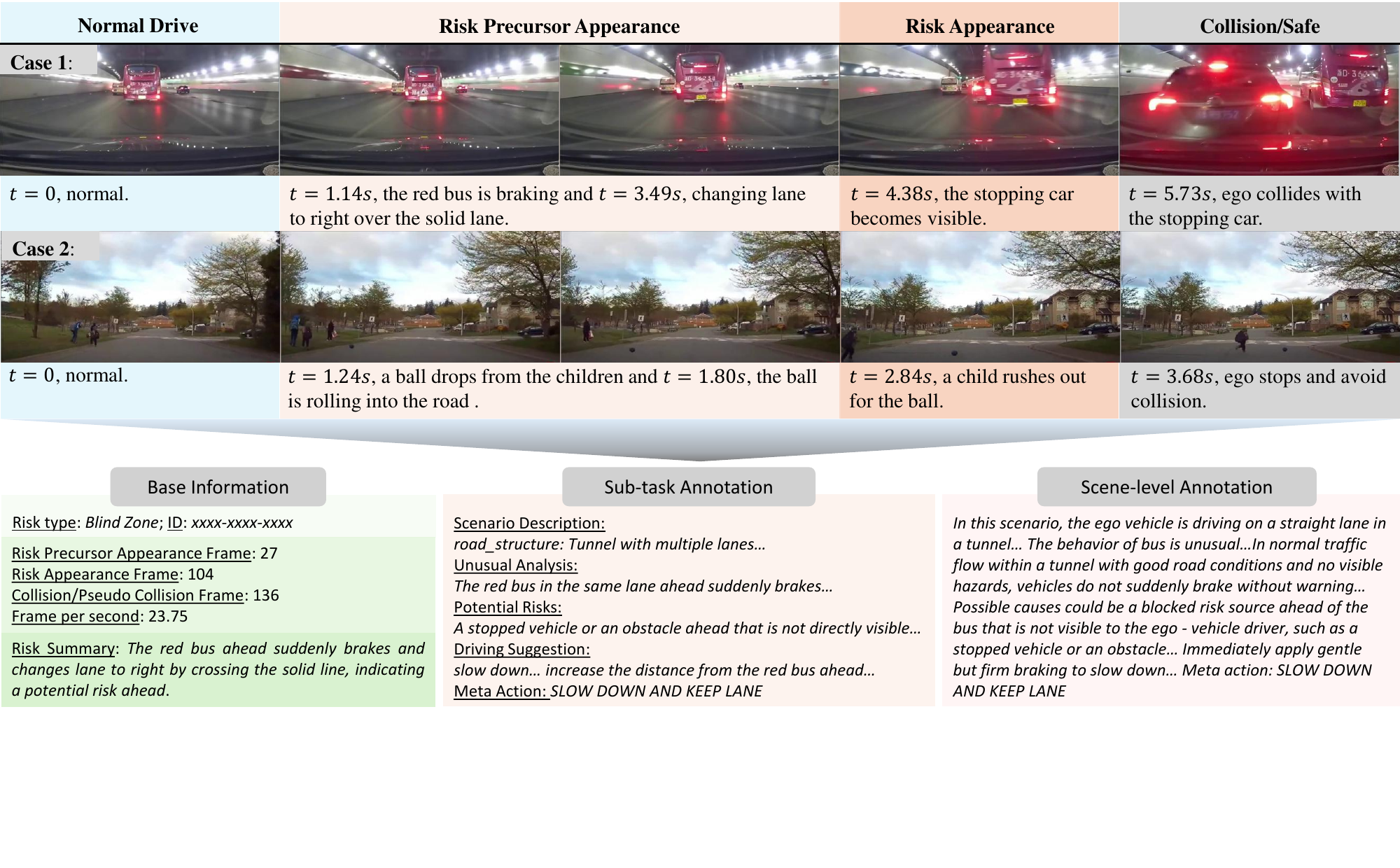}
    \caption{Temporal structure and examples of \dataset. The precursor is observable before the associated hazard becomes directly visible. Models receive only pre-risk observations; later frames support annotation of the event and its relation to the precursor.}
    \label{fig:datasetstructure}
    \vspace{-0.4cm}
\end{figure*}

An autonomous vehicle may observe a leading vehicle brake before it can see the obstacle that prompted the braking. Waiting for the obstacle to become directly visible can leave little time to respond. We study \textbf{potential-risk reasoning}: inferring a dangerous event or condition that is not yet directly observable from a visible \textbf{precursor} and the surrounding scene context \cite{lattner2005knowledge,takahashi2016explaining}. As illustrated in Fig.~\ref{fig:datasetstructure}, this task concerns the interval between precursor appearance and direct hazard visibility, before a collision or near-collision.

Interpreting a precursor requires more than recognizing an unusual object or maneuver. Braking may indicate an occluded obstacle, a traffic signal, or normal car-following behavior. The model must ground the cue in the scene, infer a plausible hazard, and recommend a proportionate response. This creates two requirements for supervision: the observation must precede direct risk visibility, and the annotation must explain how the available evidence supports the inferred risk. Without the first, evaluation may reward hazard recognition rather than anticipation; without the second, it is difficult to distinguish grounded reasoning from a plausible guess.

Existing driving datasets provide accident videos, risk descriptions, and increasingly rich interaction or planning annotations \cite{fang2021dada,malla2023drama,chen2025automated,xu2025wode2e,li2024womdreasoning}. However, these resources do not generally organize supervision around both an explicit precursor-to-risk temporal boundary and a chain linking the precursor to the hidden hazard and an appropriate response. Driving vision-language models (VLMs) offer a means of expressing this chain, but a generic reasoning prompt alone does not establish which visual evidence justifies a risk claim.

We introduce \dataset, an ego-view driving dataset comprising 523 scenarios and 3,138 scene-level and step-level question--answer pairs. Temporal annotations mark precursor appearance, direct hazard visibility, and collision or near-collision, with a median precursor-to-risk interval of 1.63\,s. Model-assisted annotations grounded in expert-established risk summaries link scene context, precursor, risk, driving suggestion, and meta-action. Experts audit a subset of these annotations. Building on this five-step structure, \method\ fine-tunes VLMs through joint scene-level and auxiliary supervision to connect observable evidence to potential risks and driving responses. The framework preserves the backbone architecture and provides high-level risk information and action suggestions for an autonomous driving system (ADS).

Our contributions are threefold:
\begin{enumerate}
    \item We introduce \dataset, a potential-risk reasoning benchmark that explicitly marks the interval from precursor appearance to direct hazard visibility, with structured annotations for training and evaluation on pre-risk observations.
    \item We propose \method, which couples a precursor-centered reasoning chain with step-level supervision. Joint scene-level and auxiliary training teaches models to connect visual evidence to potential risks and driving responses.
    \item We demonstrate the value of precursor-centered supervision for risk reasoning and downstream planning through experiments across VLM backbones, auxiliary-task ablations, and NAVSIM evaluation.
\end{enumerate}

\section{Related Work}
\subsection{Potential-risk reasoning methods}
Early potential-risk reasoning methods associate observed traffic conditions with hidden hazards through predefined knowledge, such as the possibility of a child emerging from behind a stopped bus \cite{lattner2005knowledge,takahashi2016explaining}. Knowledge-graph-based methods extend this approach by representing traffic entities and their semantic relations \cite{li2023graph}. Although these representations expose the basis of an inference, extending their coverage requires encoding additional scenarios and relations. Recent work trains VLMs to infer planning-critical occluded agents from visual evidence, providing an alternative to predefined risk patterns \cite{chahe2026hidden}. \method\ learns precursor-to-risk associations across occlusion, behavior, and object-related scenarios, with supervision explicitly linking the available evidence to a hazard before it becomes directly observable.

\subsection{Risk-centric driving datasets}
Driving risk datasets differ in viewpoint, temporal coverage, and supervision. Roadside accident collections and recordings of third-party incidents support accident understanding, but do not necessarily capture the ego vehicle's opportunity to anticipate a hazard \cite{shah2018cadp,yao2022dota}. DADA-2000, DRAMA, and CODA-LM provide incident, risk, and language annotations \cite{fang2021dada,malla2023drama,chen2025automated}, while AUTOPILOT-VQA extends incident-centered evaluation to contextual and avoidability questions \cite{damodharan2026autopilot}. Resources such as WOD-E2E broaden long-tail driving coverage \cite{xu2025wode2e}, and K-Risk pairs high-risk trajectories with language annotations, including causal analyses and action recommendations for a subset \cite{huang2026krisk}. Long-tail coverage concerns scenario frequency, whereas potential-risk reasoning concerns the timing of inference relative to direct hazard visibility. \dataset\ captures this complementary temporal dimension through precursor and risk appearance times and explicit evidence-to-risk annotations.

\subsection{Language reasoning for driving}
VLM-based driving systems use language to connect scene understanding with actions or trajectories \cite{tian2024drivevlm,wang2025alpamayo}. Think-Driver structures it from scene interpretation to decision-making \cite{zhang2024thinkdriver}, while VLM-AutoDrive uses question answering and chain-of-thought supervision for collision and near-collision detection \cite{bhat2026vlmautodrive}. Language formulations for potential-risk reasoning remain underdeveloped, motivating \method's explicit supervision of the relation between an observable precursor, a not-yet-visible hazard, and a justified response.

\section{Methodology}
\label{sec:method}

\subsection{Overview of PR-Reasoner}

% \begin{figure*}[ht]
%     \centering
%     \includegraphics[trim=2 228 195 2, clip, width=0.95\linewidth]{figures/framework.pdf}
%     \caption{Overview of \method. Five ego-view frames and a task prompt enter a vision-language backbone, which generates a chain from scene description and visible precursor to inferred risk, driving suggestion, and ADS-compatible meta-action.}
%     \label{fig:framework}
% \end{figure*}

\begin{figure}[ht]
    \centering
    \includegraphics[trim=8 60 390 80, clip, width=\linewidth]{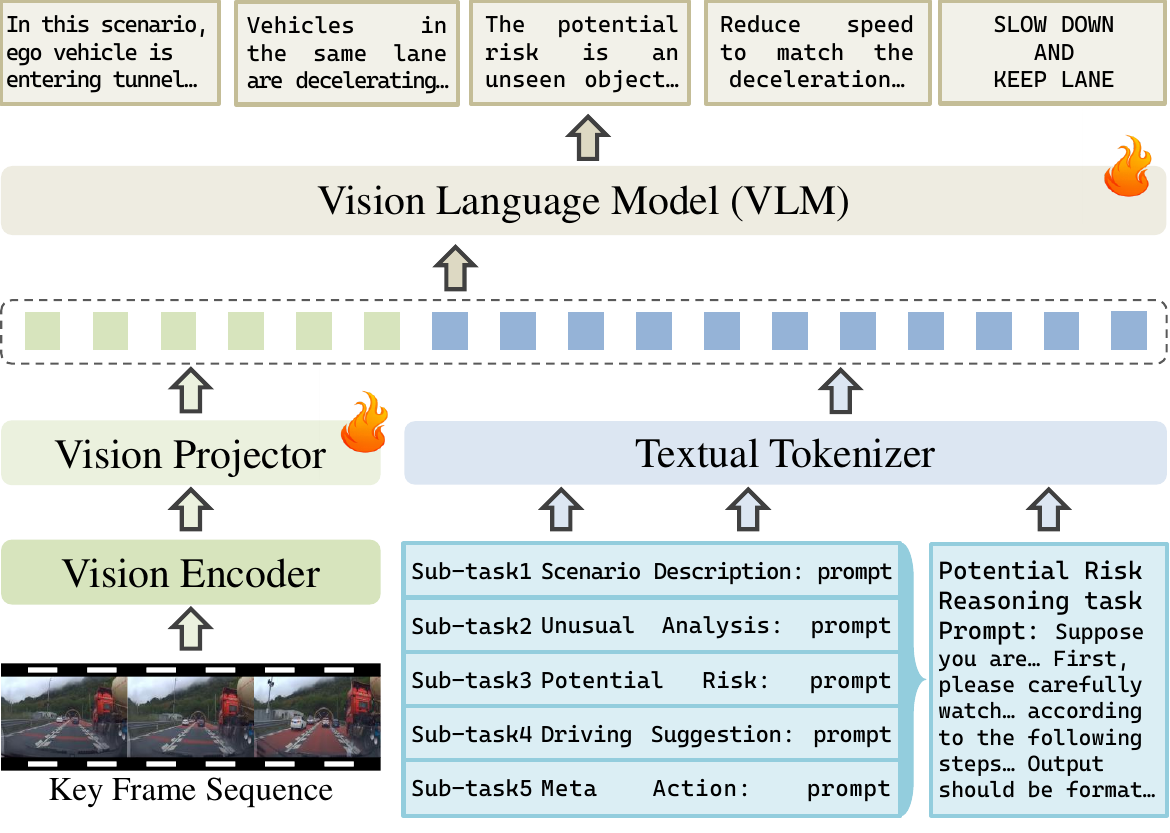}
    \caption{Overview of \method. Five ego-view frames and a task prompt enter a vision-language backbone, which generates a chain from scene description and visible precursor to inferred risk, driving suggestion, and ADS-compatible meta-action.}
    \label{fig:framework}
    \vspace{-0.3cm}
\end{figure}

\method\ maps a sequence of ego-view observations $V_{t-N:t}$ and a task prompt $T$ to a structured reasoning chain. At the evaluation time $t$, the sequence contains a visible precursor but ends before the associated hazard becomes directly observable. The model must use the current scene and short motion history to infer that hazard and specify a response. We write this mapping as
\begin{equation}
\mathcal{C}=f_{\theta}(V_{t-N:t},T)=(s,p,r,a_1,a_2),
\end{equation}

where $s$ describes the scene, $p$ identifies the visible precursor, $r$ states the potential risk, $a_1$ gives a driving suggestion, and $a_2$ specifies an ADS-compatible meta-action. This ordering makes the intermediate inference explicit. For example, a leading vehicle's brake lights can support an inferred obstacle beyond the current view, which in turn motivates slowing down while maintaining lane. Fig.~\ref{fig:framework} shows the model inputs and five supervised outputs.

We sample five frames at 0.5-s intervals over the 2-s window ending at $t$ to capture the temporal evolution of the precursor. Each backbone processes these frames and the task prompt through its native visual and text encoders. We evaluate alternative temporal windows in Section~\ref{sec:timeablation}.

\subsection{Precursor-centered reasoning chain with auxiliary sub-tasks}
The five sub-tasks follow the order of the reasoning chain. \textit{Scenario Description} ($s$) establishes the visible context, including road conditions, traffic participants, their motion, and cues such as warning lights. \textit{Unusual Analysis} ($p$) identifies the precursor and explains why it is informative in that context. The annotation can describe an abnormal maneuver, an occlusion, or a conflict between an observed behavior and the apparent road situation. \textit{Potential Risk} ($r$) states the possible hidden event, the participants involved, its likely location, and the causal link to the precursor.

The two action steps translate this inference into a driving response. \textit{Driving Suggestion} ($a_1$) describes a mitigation strategy in natural language. \textit{Meta Action} ($a_2$) specifies immediate longitudinal and lateral behavior, such as slowing down while keeping lane, together with behavior-level guidance such as avoiding overtaking. These are high-level commands for an ADS, rather than actuator controls or trajectories. Supervising the five steps individually provides explicit targets for grounding the risk inference and its resulting action.

\subsection{Training and inference}
The full training pipeline first adapts the backbone to driving-scene semantics using CODA-LM \cite{chen2025automated} and to multi-frame understanding using Lingo-QA \cite{marcu2024lingoqa}, then fine-tunes it on \dataset. This final stage combines complete scene-level answers with five auxiliary question--answer pairs per scenario. At inference, a scene-agnostic prompt requests the complete chain in a single response from the pre-risk frames; expert risk summaries and future frames are excluded. Training configurations are given in Section~\ref{sec:experimentset}, and the contribution of auxiliary supervision is evaluated in Section~\ref{sec:trainingcontrols}.

\section{\dataset\ Dataset}

\subsection{Dataset overview}
\dataset\ comprises ego-view driving scenarios from public videos, existing datasets, and controlled recordings. Each scenario includes an ID, risk category, expert risk summary, and three temporal markers (Fig.~\ref{fig:datasetstructure}): the \textbf{Precursor Appearance Frame}, when a cue first supports inference about the subsequent hazard; the \textbf{Risk Appearance Frame}, when the hazard becomes directly visible; and the \textbf{Collision or Near-Collision Frame}, marking impact or the critical point at which a collision would have occurred without evasive action. Experts verify the precursor--risk relation in the full scene context. Evaluation inputs precede risk appearance, while the subsequent event informs annotation and scoring.

Each scenario also provides one scene-level answer covering the complete reasoning chain $\mathcal{C}$ and five aligned sub-task answers, one per reasoning step. These annotations support scene-level training with or without auxiliary step-level supervision.

\subsection{Dataset statistics}
\label{sec:datastat}

\begin{figure*}[th]
    \centering
    \includegraphics[trim=55 300 61 291, clip, width=0.95\linewidth]{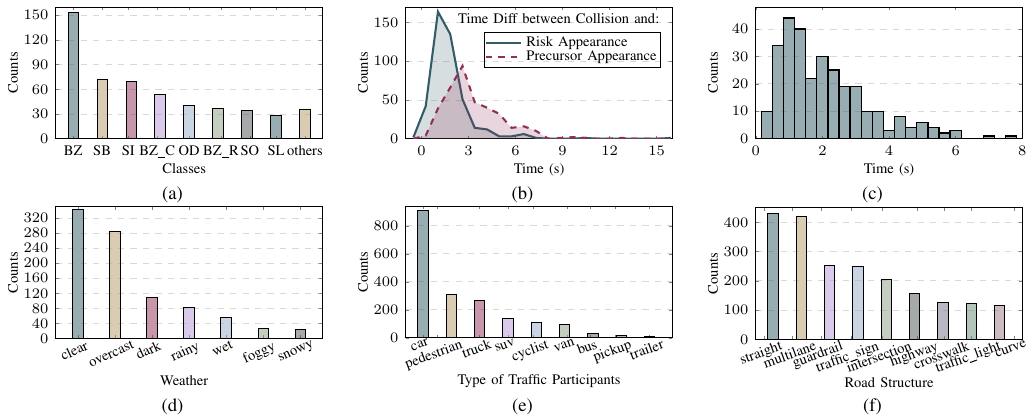}
    \caption{Composition and temporal structure of \dataset. (a) Counts of ten risk categories. (b) Intervals from precursor and direct risk appearance to collision or near-collision. (c) Time between precursor and direct risk appearance. (d--f) Distributions of weather, traffic participants, and road structures.}
    \label{fig:datasetstatistics}
    \vspace{-0.4cm}
\end{figure*}

The evaluated release contains 523 scenarios and nearly 100,000 video frames. Each scenario contributes one scene-level and five sub-task question--answer pairs, giving 3,138 pairs in total. The clips span different weather, road structures, and participant types, as in  Fig.~\ref{fig:datasetstatistics}.

Ten categories summarize recurring short-horizon risk patterns. The largest, \textit{Blind Zone (BZ)}, contains 153 scenarios in which a participant is occluded but its possible presence is suggested by road context or another participant's reaction. \textit{Special Information (SI)} contains 75 scenarios with visible signals that imply a hazard, and \textit{Sudden Behavior (SB)} contains 72 scenarios involving an impending change in motion. The other groups are \textit{Blind Zone at Crosswalk or Intersection (BZ\_C)}, \textit{Blind Zone with Right of Way (BZ\_R)}, \textit{Object Drop (OD)}, \textit{Special Objects (SO)}, \textit{Sliding (SL)}, \textit{Low Vision (LV)}, and \textit{Road Surface Abnormal (RS)}. They cover occlusion at crossings, unexpected yielding, cargo loss, unusual objects, loss of traction, degraded visibility, and inferred surface defects. When multiple patterns apply, experts select the one that best explains the main risk.

The median interval from precursor appearance to collision or near-collision is approximately 2.9\,s, compared with 1.5\,s from direct risk appearance to the same endpoint (Fig.~\ref{fig:datasetstatistics}b). The precursor-to-risk interval has a median of 1.63\,s (Fig.~\ref{fig:datasetstatistics}c), quantifying the temporal opportunity for anticipation before the hazard becomes directly visible.

\subsection{Dataset construction}
\label{sec:datasetconstruction}
\subsubsection{Data collection}
We curated first-person driving videos from YouTube and Bilibili, existing datasets, and controlled real-vehicle recordings. Dataset sources include DrivingDojo \cite{wang2024drivingdojo}, MM-AU \cite{Fang2024mmau}, and DADA-2000 \cite{fang2021dada}. Available event labels and descriptions supported initial retrieval. For example, we searched occlusion-related DrivingDojo categories and accident-cause descriptions in MM-AU. Experts then examined the candidate videos and retained scenarios for which a visible precursor could reasonably support inference about the subsequent hazard from the ego view.

% Controlled recordings supplement the public sources with representative patterns observed during curation. We reproduced selected situations in a test field while varying participant type, relative position, and motion. A front-facing onboard camera recorded these cases at 1080p resolution with a $121^\circ$ field of view and a 2.8\,mm focal length. We converted all retained clips to a common format before selecting key frames and preparing annotations.

\subsubsection{Model-assisted annotation and expert review}

For each scenario, two experts jointly establish a risk summary, identifying the visible precursor, subsequent hazard, and their causal relation. The annotation models then receive the key-frame sequence, task-specific prompts, and expert risk summary, together with preceding sub-task outputs when annotating successive reasoning steps. They generate the five sub-task annotations in order, followed by the complete scene-level answer.

The annotation pipeline comprises draft generation, independent checking, revision merging, and expert spot-checking. Following prior multi-agent annotation practice \cite{huang2025drivesotif}, we use Doubao \cite{doubao2023}, Qwen-VL \cite{bai2025qwen3}, and GPT-4o \cite{hurst2024gpt} for the first three stages to reduce dependence on a single annotation model. For each sub-task, a generation agent drafts an answer, and an independent checking agent examines the video for omissions and unsupported claims. A merging agent verifies the proposed revisions against the video, incorporates valid corrections, and standardizes the output. Scene-level answers conduct the same model-assisted stages, with sub-task annotations available during checking.

In the final stage, experts audited 10\% of the resulting annotations, rating each reasoning step, cross-step consistency, and avoidance of unsupported content on four ordinal levels. Normalized scores ranged from 0.874 for avoidance of unsupported content to 0.943 for scene description; potential-risk annotations scored 0.887. Only 3.8\% of audited samples required re-annotation.

\section{Experiments}
\label{sec:experiments}

\subsection{Experimental setup}
\label{sec:experimentset}

We evaluate seven VLM backbones with 1B to 8B parameters: Qwen3-VL-2B/4B/8B \cite{bai2025qwen3} and InternVL3.5-1B/2B/4B/8B \cite{chen2024internvl}. We evaluate complete scene-level responses, using the same precursor-centered prompt for each pretrained backbone and its \method\ counterpart.
We partition the 523 scenarios into training, validation, and test sets at a 75:10:15 ratio, keeping all question--answer pairs from a scenario in the same split.

We use LoRA-SFT in ms-swift \cite{zhao2024swiftascalablelightweightinfrastructure} for four epochs on NVIDIA H20 GPUs with 96\,GB memory. The learning rate is $10^{-4}$, LoRA rank is 8, alpha is 32, warmup ratio is 0.05, and gradient accumulation spans eight steps. We use bf16 precision and per-GPU batch sizes of 2 to 16 according to model size. Images have a longest edge of 480 pixels. We select the checkpoint with the lowest validation causal loss. Inference uses greedy decoding for deterministic results. Latency experiments use batch size one on the same hardware.
The controlled experiments use InternVL3.5-4B (In-4B) and Qwen3-VL-8B (Qw-8B); category diagnostics and NAVSIM evaluation use Qw-8B only.

% Table generated from avg_scores_ci.json
\begin{table*}[thbp]
  \centering
  \setlength{\tabcolsep}{3pt}
  \caption{Potential-risk reasoning across seven VLM backbones. Baseline denotes the original model, and PR-Reasoner denotes driving-domain pretraining followed by PotentialRiskQA fine-tuning. InternVL3.5 and Qwen3-VL are abbreviated In and Qw. Values are test-scenario means; $\pm$ gives the larger distance to either bound of the 95\% bootstrap confidence interval, obtained by resampling test scenarios with replacement 10,000 times and recomputing the mean. All metrics are higher-is-better. Bold marks the highest mean within each setting and metric.}
    \begin{tabular}{ccccccc}
    \hline
    \multicolumn{1}{c}{Group} & VLMs & \multicolumn{1}{c}{Format} & \multicolumn{1}{c}{SceUnd} & \multicolumn{1}{c}{RiskCog} & \multicolumn{1}{c}{FalseRisk} & \multicolumn{1}{c}{Action} \\
     \hline
    \multirow{7}[0]{*}{\textbf{Baseline}} & In-1B & 0.774{\scriptsize$\,\pm\!$0.034} & 0.244{\scriptsize$\,\pm\!$0.037} & 0.256{\scriptsize$\,\pm\!$0.073} & 0.051{\scriptsize$\,\pm\!$0.038} & 0.327{\scriptsize$\,\pm\!$0.038} \\
     & In-2B & 0.641{\scriptsize$\,\pm\!$0.047} & 0.187{\scriptsize$\,\pm\!$0.035} & 0.265{\scriptsize$\,\pm\!$0.081} & 0.094{\scriptsize$\,\pm\!$0.056} & 0.274{\scriptsize$\,\pm\!$0.030} \\
     & In-4B & 0.662{\scriptsize$\,\pm\!$0.051} & 0.250{\scriptsize$\,\pm\!$0.040} & 0.291{\scriptsize$\,\pm\!$0.085} & 0.098{\scriptsize$\,\pm\!$0.060} & 0.504{\scriptsize$\,\pm\!$0.042} \\
     & In-8B & 0.829{\scriptsize$\,\pm\!$0.043} & 0.297{\scriptsize$\,\pm\!$0.041} & 0.321{\scriptsize$\,\pm\!$0.090} & \textbf{0.128}{\scriptsize$\,\pm\!$0.073} & 0.359{\scriptsize$\,\pm\!$0.044} \\
     & Qw-2B & \textbf{0.842}{\scriptsize$\,\pm\!$0.047} & 0.226{\scriptsize$\,\pm\!$0.044} & 0.251{\scriptsize$\,\pm\!$0.082} & 0.017{\scriptsize$\,\pm\!$0.017} & 0.475{\scriptsize$\,\pm\!$0.045} \\
     & Qw-4B & 0.769{\scriptsize$\,\pm\!$0.060} & 0.304{\scriptsize$\,\pm\!$0.047} & 0.222{\scriptsize$\,\pm\!$0.085} & 0.094{\scriptsize$\,\pm\!$0.064} & 0.414{\scriptsize$\,\pm\!$0.038} \\
     & Qw-8B & 0.761{\scriptsize$\,\pm\!$0.051} & \textbf{0.349}{\scriptsize$\,\pm\!$0.044} & \textbf{0.359}{\scriptsize$\,\pm\!$0.094} & 0.043{\scriptsize$\,\pm\!$0.043} & \textbf{0.595}{\scriptsize$\,\pm\!$0.049} \\
          \hline
    \multirow{7}[0]{2.1cm}{\centering\textbf{PR-Reasoner}} & In-1B & 0.915{\scriptsize$\,\pm\!$0.034} & 0.359{\scriptsize$\,\pm\!$0.045} & 0.274{\scriptsize$\,\pm\!$0.090} & 0.030{\scriptsize$\,\pm\!$0.038} & 0.789{\scriptsize$\,\pm\!$0.030} \\
     & In-2B & 0.833{\scriptsize$\,\pm\!$0.038} & 0.377{\scriptsize$\,\pm\!$0.040} & \textbf{0.449}{\scriptsize$\,\pm\!$0.098} & 0.158{\scriptsize$\,\pm\!$0.077} & 0.813{\scriptsize$\,\pm\!$0.038} \\
     & In-4B & 0.927{\scriptsize$\,\pm\!$0.034} & 0.401{\scriptsize$\,\pm\!$0.036} & 0.368{\scriptsize$\,\pm\!$0.094} & 0.162{\scriptsize$\,\pm\!$0.077} & \textbf{0.832}{\scriptsize$\,\pm\!$0.032} \\
     & In-8B & 0.953{\scriptsize$\,\pm\!$0.026} & 0.431{\scriptsize$\,\pm\!$0.047} & 0.419{\scriptsize$\,\pm\!$0.098} & 0.167{\scriptsize$\,\pm\!$0.077} & 0.803{\scriptsize$\,\pm\!$0.032} \\
     & Qw-2B & 0.919{\scriptsize$\,\pm\!$0.064} & 0.407{\scriptsize$\,\pm\!$0.057} & 0.338{\scriptsize$\,\pm\!$0.097} & 0.167{\scriptsize$\,\pm\!$0.083} & 0.792{\scriptsize$\,\pm\!$0.031} \\
     & Qw-4B & 0.906{\scriptsize$\,\pm\!$0.034} & 0.439{\scriptsize$\,\pm\!$0.048} & 0.389{\scriptsize$\,\pm\!$0.098} & 0.137{\scriptsize$\,\pm\!$0.073} & 0.774{\scriptsize$\,\pm\!$0.032} \\
     & Qw-8B & \textbf{0.957}{\scriptsize$\,\pm\!$0.026} & \textbf{0.483}{\scriptsize$\,\pm\!$0.050} & 0.389{\scriptsize$\,\pm\!$0.098} & \textbf{0.261}{\scriptsize$\,\pm\!$0.098} & 0.807{\scriptsize$\,\pm\!$0.029} \\
          \hline
    \end{tabular}%
      \label{tab:mainresults}%

      \vspace{-10pt}
\end{table*}%

\subsection{Evaluation metrics}

We assess responses using five metrics. Format is computed by rules; the remaining metrics use Qwen-VL-max as a judge, following language-based driving evaluation practice \cite{chen2025automated,wang2023large,lu2023llmscore}. The judge receives the candidate response, image sequence, ground-truth annotations, and expert-identified accident reason, with the same reference information and scoring prompts used for every model.

\textbf{Format} checks whether the response contains the required reasoning steps and follows the meta-action structure. An unparseable response scores zero. Otherwise, its raw score is
\begin{equation}
F_i=(3-H_i-C_i)\exp[-0.0005\max(0,L_i-2000)],
\end{equation}
where $L_i$ is the response length in characters, and $H_i$ and $C_i$ indicate a human-driver instruction in the driving suggestion and a conditional expression in the meta-action, respectively. Each violation deducts one point, and responses exceeding 2,000 characters receive a length penalty.

\textbf{Scene Understanding (SceUnd)} assesses three components: key-object coverage, risk-relevant scene information, and avoidance of unsupported scene details. Their scores are combined as a weighted mean with weights $1$, $2$, and $1$, respectively.

\textbf{Risk Cognition (RiskCog)} assesses whether the inferred hazard, involved participants, and precursor-to-risk explanation agree with the expert reference.

\textbf{False Risk (FalseRisk)} assesses whether risk claims are supported by the scene and reference annotations, penalizing invented participants, unsupported behavior, and implausible hazards. Higher scores indicate fewer unsupported risk claims.

\textbf{Driving Action (Action)} assesses safety, efficiency, and ADS executability, using a weighted mean with weights $2$, $1$, and $2$, respectively. Suggested actions should address the inferred hazard without unnecessary braking or stopping and provide concrete, consistent commands.

RiskCog, FalseRisk, and each of the three components of SceUnd and Action use the same four-level rubric: 0 (completely incorrect), 1 (weakly correct), 2 (mostly correct), and 3 (fully correct). The judge scores each criterion and provides a brief justification.
Finally, we normalize all five metrics to $[0,1]$. We report the mean normalized score over evaluation scenarios; higher is better for every metric.

\subsection{Main benchmark results}

\method\ improves mean scene understanding, risk cognition, and action quality across all seven backbones (Table~\ref{tab:mainresults}). Action shows the largest absolute gain for every backbone, indicating that the full training pipeline is particularly effective at producing driving responses that satisfy the safety, efficiency, and executability criteria. Improved Format scores accompany these content gains, while FalseRisk improves on all backbones except In-1B. The remaining unsupported risk claims show that learning the response structure and action format does not fully resolve the difficulty of grounding a hidden hazard in visual evidence.

Model size affects the reasoning stages differently. After training, SceUnd increases monotonically with size within both families, but RiskCog and Action do not: In-2B achieves the highest RiskCog, whereas In-4B achieves the highest Action. Larger backbones therefore provide stronger scene descriptions without uniformly improving the subsequent inference and response. The following controls examine the contributions of prompting and supervision to these gains.

\subsection{Prompt and training controls}
\label{sec:trainingcontrols}

\subsubsection{Prompting versus precursor supervision}
\begin{table}[ht]
\centering
\caption{Prompt controls for In-4B and Qw-8B on identical pre-risk frames. Higher scores indicate better performance.}
\label{tab:prompt_control}
\setlength{\tabcolsep}{3pt}
\begin{tabular}{llccc}
\hline
Backbone & Setting & RiskCog & FalseRisk & Action \\
\hline
\multirow{4}{*}{In-4B} & Direct risk CoT & 0.269 & 0.098 & 0.444 \\
 & Think-Driver CoT & 0.329 & \textbf{0.175} & 0.290 \\
 & PR prompt & 0.291 & 0.098 & 0.504 \\
 & PR-Reasoner & \textbf{0.368} & 0.162 & \textbf{0.832} \\
\hline
\multirow{4}{*}{Qw-8B} & Direct risk CoT & 0.338 & 0.184 & 0.577 \\
 & Think-Driver CoT & 0.346 & 0.139 & 0.549 \\
 & PR prompt & 0.359 & 0.043 & 0.595 \\
 & PR-Reasoner & \textbf{0.389} & \textbf{0.261} & \textbf{0.807} \\
\hline
\end{tabular}
      \vspace{-10pt}

\end{table}

We compare three reasoning prompts using identical backbones and visual inputs (Table~\ref{tab:prompt_control}). \textit{Direct risk CoT} directly predicts the risk and then proposes a driving response ($r\rightarrow a$). \textit{Think-Driver CoT}, adapted from \cite{zhang2024thinkdriver}, adds scene understanding before risk prediction ($s\rightarrow r\rightarrow a$). The precursor-centered PR prompt further introduces explicit precursor identification ($s\rightarrow p\rightarrow r\rightarrow a$). 

Relative to Direct risk CoT, Think-Driver CoT improves RiskCog on both backbones but lowers Action. The PR prompt improves both RiskCog and Action over Direct risk CoT, although its FalseRisk is unchanged on In-4B and decreases on Qw-8B. Explicitly asking for precursor reasoning therefore helps connect risk analysis to a response, but prompting alone does not reliably constrain unsupported hazard claims.
Further-trained \method\ models improve all three reported content metrics over their prompt-only counterparts. The gains thus extend beyond prompt design. This comparison reflects the complete training pipeline; the following ablation examines the contribution of auxiliary supervision.

\begin{table*}[t]
\centering
\caption{Effect of auxiliary supervision on In-4B and Qw-8B. Baseline is the original VLM. ``Scene only'' uses complete scene-level answers, whereas ``Scene + auxiliary'' also uses five step-level answers. Overall metrics and their components are reported together for each model and supervision setting. All scores are higher-is-better.}
\label{tab:ablation}
\setlength{\tabcolsep}{3.5pt}
\renewcommand{\arraystretch}{1.1}
\begin{tabular}{llccccccccccc}
\hline
\multirow{2}{*}{Model} & \multirow{2}{*}{Supervision} & \multirow{2}{*}{Format} & \multirow{2}{*}{SceUnd} & \multirow{2}{*}{RiskCog} & \multirow{2}{*}{FalseRisk} & \multirow{2}{*}{Action} & \multicolumn{3}{c}{Scene understanding} & \multicolumn{3}{c}{Driving action} \\
\cline{8-13}
 & & & & & & & Key objects & Risk info. & No halluc. & Safety & Efficiency & Executability \\
\hline
In-4B & Baseline & 0.662 & 0.250 & 0.291 & 0.098 & 0.504 & 0.470 & 0.128 & 0.274 & 0.423 & 0.701 & 0.487 \\
In-4B & Scene only & \textbf{0.927} & 0.350 & 0.299 & 0.068 & 0.803 & 0.560 & 0.286 & 0.269 & 0.744 & 0.808 & 0.859 \\
In-4B & Scene + auxiliary & \textbf{0.927} & \textbf{0.401} & \textbf{0.368} & \textbf{0.162} & \textbf{0.832} & \textbf{0.603} & \textbf{0.321} & \textbf{0.359} & \textbf{0.752} & \textbf{0.876} & \textbf{0.889} \\
\hline
Qw-8B & Baseline & 0.761 & 0.349 & 0.359 & 0.043 & 0.595 & 0.568 & 0.274 & 0.282 & 0.547 & 0.658 & 0.611 \\
Qw-8B & Scene only & \textbf{0.983} & 0.417 & 0.376 & 0.098 & 0.721 & 0.632 & 0.363 & 0.308 & 0.645 & 0.761 & 0.778 \\
Qw-8B & Scene + auxiliary & 0.957 & \textbf{0.483} & \textbf{0.389} & \textbf{0.261} & \textbf{0.807} & \textbf{0.641} & \textbf{0.423} & \textbf{0.444} & \textbf{0.731} & \textbf{0.838} & \textbf{0.868} \\
\hline
\end{tabular}
      \vspace{-10pt}

\end{table*}

\subsubsection{Contribution of driving-domain pretraining}
We compare training with and without driving dataset pretraining, keeping scene-level and auxiliary \dataset\ supervision identical for each backbone (Table~\ref{tab:pretrainablation}).

\dataset\ fine-tuning alone improves SceUnd, RiskCog, and Action over the original models on both backbones (Tables~\ref{tab:pretrainablation} and \ref{tab:mainresults}). These gains establish a contribution from the task-specific supervision independently of the additional driving-domain data. FalseRisk responds differently across backbones, improving on Qw-8B but declining on In-4B. 
Adding driving-domain pretraining further improves SceUnd, FalseRisk, and Action on both backbones. Its benefit is therefore consistent for scene grounding, avoidance of unsupported risk claims, and action quality. RiskCog, however, improves on In-4B and declines on Qw-8B. Broader driving supervision complements \dataset\ training, but better scene understanding does not necessarily translate into more accurate inference of the specific hidden hazard.

\begin{table}[t]
\centering
\caption{Effect of driving-domain pretraining on InternVL3.5-4B (In-4B) and Qwen3-VL-8B (Qw-8B). All settings receive the same PotentialRiskQA fine-tuning with scene-level and auxiliary tasks. ``Yes'' adds CODA-LM and Lingo-QA pretraining. Higher scores are better.}
\label{tab:pretrainablation}
\setlength{\tabcolsep}{2.5pt}
\renewcommand{\arraystretch}{1.08}
\begin{tabular}{llccccc}
\hline
Model & Pretrain & Format & SceUnd & RiskCog & FalseRisk & Action \\
\hline
\multirow{2}{*}{In-4B} & No  & 0.919 & 0.341 & 0.321 & 0.077 & 0.762 \\
                       & Yes & \textbf{0.927} & \textbf{0.401} & \textbf{0.368} & \textbf{0.162} & \textbf{0.832} \\
\hline
\multirow{2}{*}{Qw-8B} & No  & \textbf{0.962} & 0.399 & \textbf{0.416} & 0.147 & 0.795 \\
                       & Yes & 0.957 & \textbf{0.483} & 0.389 & \textbf{0.261} & \textbf{0.807} \\
\hline
\end{tabular}
\vspace{-10pt}
\end{table}

\subsubsection{Contribution of auxiliary sub-tasks}

We compare scene-level supervision with and without the five auxiliary tasks, retaining driving-domain pretraining in both settings (Table~\ref{tab:ablation}). The original backbone is included as a reference. The table reports overall metrics and their scene-understanding and driving-action components side by side for each model and supervision setting.

Auxiliary supervision improves SceUnd, RiskCog, FalseRisk, and Action on both backbones, indicating that explicit targets for intermediate reasoning steps strengthen scene understanding, risk inference, avoidance of unsupported claims, and action quality beyond scene-level supervision alone. The component scores likewise improve across scene interpretation and action generation. Format remains unchanged on In-4B and decreases slightly on Qw-8B, while the content-oriented metrics improve.

\subsection{Temporal context}
\begin{table}[ht]
\centering
\setlength{\tabcolsep}{2.5pt}
\caption{Effect of temporal context on InternVL3.5-4B (In-4B) and Qwen3-VL-8B (Qw-8B). }
\begin{tabular}{cccccccc}
\hline
Window & Model & Format & SceUnd & RiskCog & FalseRisk & Action & Time (s) \\
\hline
\multirow{2}{*}{1\,s}
 & In-4B & \textbf{0.932} & 0.359 & \textbf{0.393} & 0.064 & 0.823 & \textbf{0.99} \\
 & Qw-8B & 0.863 & 0.402 & \textbf{0.444} & 0.107 & 0.809 & 1.64 \\
\hline
\multirow{2}{*}{2\,s}
 & In-4B & 0.927 & \textbf{0.401} & 0.368 & \textbf{0.162} & \textbf{0.832} & 1.49 \\
 & Qw-8B & 0.957 & \textbf{0.483} & 0.389 & \textbf{0.261} & 0.807 & 2.35 \\
\hline
\multirow{2}{*}{3\,s}
 & In-4B & 0.893 & 0.363 & 0.338 & 0.081 & 0.803 & 1.97 \\
 & Qw-8B & \textbf{0.966} & 0.447 & 0.363 & 0.103 & \textbf{0.815} & 3.15 \\
\hline
\end{tabular}
      \vspace{-10pt}

\label{tab:window_length}
\end{table}

\label{sec:timeablation}

We vary the observation window from 1 to 3\,s at a fixed sampling interval, yielding three, five, or seven input frames (Table~\ref{tab:window_length}). The shortest window gives the highest RiskCog on both backbones but lower FalseRisk, revealing a trade-off between matching the reference hazard and avoiding additional unsupported claims. Extending the window to 2\,s gives the best scene understanding and false-risk avoidance for both models.

A further extension to 3\,s decreases SceUnd and FalseRisk while increasing latency, whereas Action changes little. Most precursor-to-risk intervals in \dataset\ occur within 2\,s and thus one possible explanation is that the additional history contains information less relevant to the annotated precursor-risk relation. This extra context may dilute the evidence needed for precursor-centered inference. We retain the 2-s window because it offers the strongest combination of scene grounding and false-risk avoidance among the tested settings.

\subsection{Category diagnostics and failure analysis}
\subsubsection{Descriptive category diagnostics}
Because the test set contains few scenarios per category, we use the full dataset of 523 scenarios, including training data, for more informative category-level diagnostics. We compare Qw-8B-based \method\ with the original Qw-8B across ten risk categories (Table~\ref{tab:category_results}); these comparisons describe performance on the full dataset rather than generalization to unseen scenarios.
On the full dataset, \method\ improves risk cognition and reduces unsupported risk claims across a broad range of scenarios. These descriptive trends suggest that precursor-centered supervision helps the model interpret indirect behavior, occlusion relationships, and unusual objects in the analyzed scenarios. Action improves across all ten categories in this comparison, including those with lower RiskCog, indicating more consistent gains in driving responses than in precise hazard identification.
\begin{table}[ht]
\centering
\caption{Risk cognition, false-risk avoidance, and action scores by category across all 523 scenarios. Raw denotes the original Qwen3-VL-8B. PR denotes PR-Reasoner with the same backbone. Category are defined in Section~\ref{sec:datastat}.}
\label{tab:category_results}
\setlength{\tabcolsep}{6pt}
\begin{tabular}{lccccccc}
\hline
Category & $N$ & \multicolumn{2}{c}{RiskCog} & \multicolumn{2}{c}{FalseRisk} & \multicolumn{2}{c}{Action} \\
 & & Raw & PR & Raw & PR & Raw & PR \\
\hline
BZ & 153 & 0.272 & 0.453 & 0.068 & 0.200 & 0.576 & 0.801 \\
SI & 75 & 0.444 & 0.396 & 0.116 & 0.111 & 0.592 & 0.799 \\
SB & 72 & 0.333 & 0.380 & 0.000 & 0.037 & 0.569 & 0.795 \\
BZ\_C & 55 & 0.406 & 0.291 & 0.061 & 0.079 & 0.559 & 0.754 \\
OD & 40 & 0.608 & 0.667 & 0.100 & 0.233 & 0.578 & 0.790 \\
BZ\_R & 37 & 0.252 & 0.315 & 0.081 & 0.054 & 0.539 & 0.762 \\
SO & 37 & 0.279 & 0.441 & 0.126 & 0.153 & 0.530 & 0.708 \\
SL & 28 & 0.583 & 0.488 & 0.143 & 0.167 & 0.657 & 0.814 \\
LV & 22 & 0.818 & 0.758 & 0.318 & 0.455 & 0.648 & 0.779 \\
RS & 4 & 0.583 & 0.000 & 0.250 & 0.250 & 0.600 & 0.767 \\
\hline
\end{tabular}
\vspace{-10pt}

\end{table}

\begin{figure*}[th]
    \centering
    \includegraphics[trim=0 175 0 0, clip, width=0.95\linewidth]{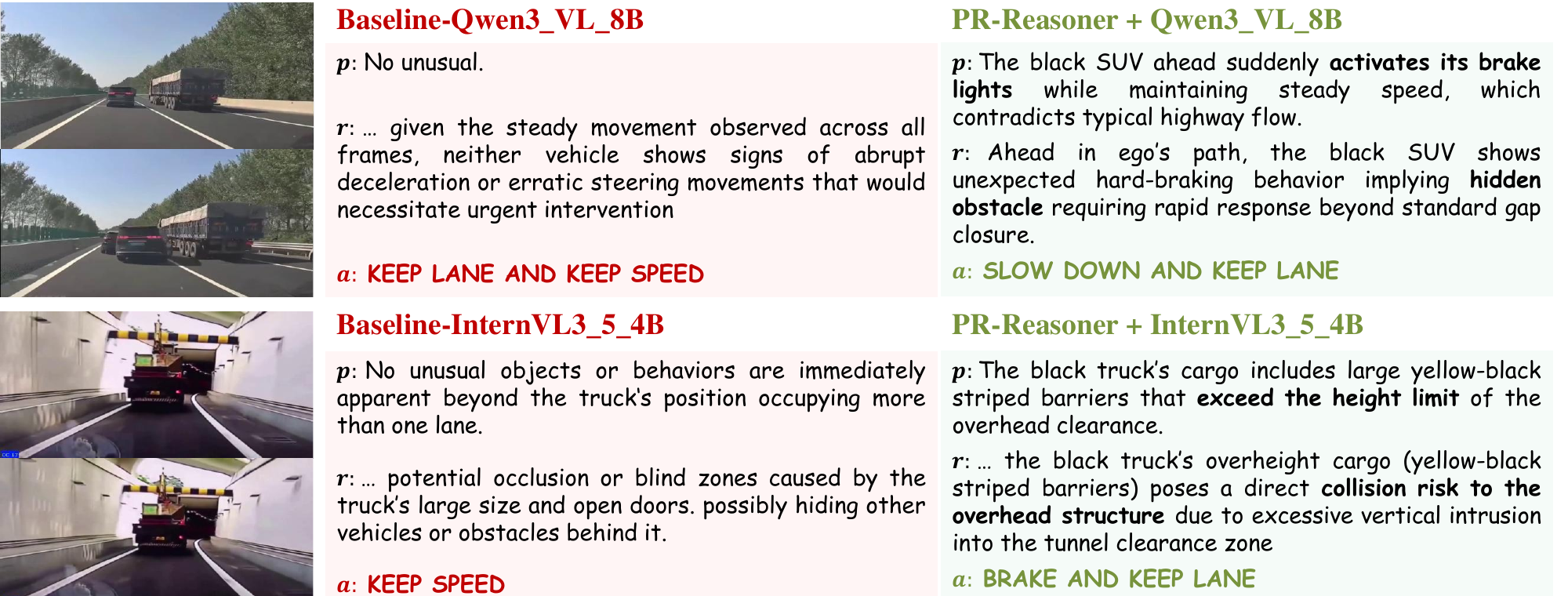}
    \caption{Precursor-grounded reasoning in two driving scenarios. The highway example tests whether the model interprets a leading SUV's brake lights as evidence of a possible hidden hazard. The tunnel example tests whether it links overheight cargo to clearance risk. Each case compares the original model with \method.}
    \label{fig:qualitiveresutls}
    \vspace{-0.4cm}
\end{figure*}

The category-level declines on the full dataset reveal two different difficulties. In special-information and right-of-way blind-zone scenarios, lower FalseRisk indicates that the model still infers hazards unsupported by the available cues, even though RiskCog improves in the latter category. In crosswalk or intersection blind zones, sliding, and low vision, RiskCog decreases despite improved FalseRisk: fewer unsupported claims do not ensure identification of the reference hazard. Error inspection of crosswalk and intersection cases reveals misjudgment of the ego vehicle's spatial approach, particularly at higher speeds. The road-surface category contains only four scenarios, limiting interpretation of its decline.

\subsubsection{Failure modes}
To examine the reasoning failures underlying these full-dataset patterns, we inspect low-scoring \method\ outputs from Qw-8B. Incorrect binding of entities, positions, and actions is the most frequent error, occurring in 80.6\% of the inspected samples. This spatial and motion grounding difficulty is consistent with the approach errors observed at crosswalks and intersections. Missed precursors and mappings from an incorrect cue to a familiar risk pattern are also common: the former can prevent identification of the reference hazard, while the latter can generate unsupported claims. Additional invented hazards occur less frequently. Among the inspected outputs, these overlapping errors identify difficulties in selecting and relating risk-relevant evidence, even when the model can describe the scene or recommend a plausible action.

\subsubsection{Precursor-grounded examples}

Two selected cases illustrate successful evidence-to-risk connections (Fig.~\ref{fig:qualitiveresutls}). In the highway scene, \method\ interprets the leading SUV's brake lights as a precursor to a possible hidden hazard and recommends slowing down, whereas the original model maintains speed. In the tunnel scene, it relates overheight cargo to a clearance conflict and recommends braking while keeping lane; the original model instead describes an unrelated risk. Both examples link a specific observation to the anticipated hazard and an appropriate response.

\subsection{Transfer to downstream planning}
\begin{table}[hbt]
\centering
\caption{Open-loop NAVSIM v1 planning results. PR-Reasoner combines PR prompt with PotentialRiskQA fine-tuning. NC no-fault collision, DAC drivable-area compliance, TTC time to collision, EP ego progress, and PDMS the overall planning score. ADE and FDE are trajectory errors. Higher is better except ADE and FDE.}
\label{tab:navsim_transfer}
\setlength{\tabcolsep}{2.2pt}
\begin{tabular}{lcccccccc}
\hline
VLM setting & NC & DAC & TTC & Comfort & EP & \textbf{PDMS} & ADE & FDE \\
\hline
Direct risk CoT & 92.26 & 78.92 & 83.62 & \textbf{100.00} & 64.75 & 67.11 & 0.844 & 2.201 \\
PR prompt & 92.90 & 78.22 & 85.35 & \textbf{100.00} & 63.61 & 67.13 & 0.842 & 2.190 \\
PR-Reasoner & \textbf{94.41} & \textbf{81.52} & \textbf{87.95} & \textbf{100.00} & \textbf{66.81} & \textbf{71.09} & \textbf{0.801} & \textbf{2.076} \\
\hline
\end{tabular}
\vspace{-10pt}
\end{table}

We next ask whether the trained risk representation benefits a separate downstream task. Using Qw-8B, we evaluate on NAVSIM v1 \cite{dauner2024navsim}. A three-layer MLP planner based on the Ego Status MLP in \cite{dauner2024navsim} separately encodes ego-state information and VLM hidden states, normalizes the features, and fuses them as $h_{\mathrm{fused}}=h_{\mathrm{ego}}+\alpha h_{\mathrm{VLM}}$, where $\alpha$ is learned. We use the first 20\% of each official split, comprising 14,624 training, 3,174 validation, and 2,424 test examples. The direct-risk and precursor prompts on the original VLM provide a prompt-level control before comparison with the trained \method\ representation.

Changing the original VLM's prompt leaves the overall planning score nearly unchanged. Under the same precursor-centered prompt, \method\ increases PDMS from 67.13 to 71.09 (Table~\ref{tab:navsim_transfer}). The improvement spans no-collision, drivable-area compliance, time-to-collision, and ego-progress scores, together with lower trajectory errors. The joint gains in safety-related scores and progress suggest that the planner's improvement does not simply come from more conservative motion. These results suggest that the representation learned through the PR-Reasoner training pipeline contains information useful for trajectory planning, extending its benefits beyond language-based evaluation.

% The trained representation thus benefits a downstream task beyond language evaluation, whereas prompt changes alone have little effect on overall planning quality. This is transfer evidence for the full training pipeline under the evaluated open-loop subset; its implications for closed-loop driving require further validation.

\subsection{Inference cost}
% !TeX root = ../RAL_main.tex
\begin{table}[ht]
\centering
\caption{End-to-end latency (s) on H20 hardware with different prompt styles. Multiples are relative to the direct meta-action baseline within each column.}
\label{tab:latency_output}
\setlength{\tabcolsep}{5pt}
\begin{tabular}{lcc}
\hline
Output setting & In-4B & Qw-8B \\
\hline
Direct meta-action & 1.15 (1.00$\times$) & 2.02 (1.00$\times$) \\
Direct risk CoT & 1.28 (1.11$\times$) & 2.15 (1.06$\times$) \\
Think-Driver-style CoT & 1.33 (1.16$\times$) & 2.11 (1.04$\times$) \\
Full PR-Reasoner chain & 1.49 (1.30$\times$) & 2.35 (1.16$\times$) \\
\hline
\end{tabular}
\vspace{-10pt}
\end{table}

On the same backbone, generating the full \method\ chain increases latency by approximately 16--30\% compared with direct meta-action output and by 9--16\% compared with Direct risk CoT or Think-Driver-style CoT (Table~\ref{tab:latency_output}).
%  The measured growth is therefore moderate relative to the cost of exposing the complete reasoning chain, while shorter outputs provide lower-latency alternatives whose reasoning and planning quality remains to be evaluated.

\section{Conclusion}
\label{sec:conclusion}

We presented \dataset\ and \method\ as a data and learning framework for connecting observable precursors to potential risks and driving responses. Temporal annotations define the evidence available before risk appearance, while the structured reasoning chain and auxiliary supervision make the intermediate inference an explicit learning target. Experiments show improved scene understanding, risk cognition, and action quality across VLM backbones, with auxiliary tasks further reducing unsupported risk claims and improving driving responses. Full-dataset category diagnostics show consistent action gains across categories. NAVSIM evaluation shows that the trained representations also benefit downstream planning. These findings support the value of explicitly supervising the precursor-to-risk relation when learning to anticipate driving hazards. Category-specific errors nevertheless reveal difficulties in relating traffic participants, spatial context, and motion to the correct risk. Future work will expand scenario coverage, strengthen this grounding, and evaluate how the resulting risk predictions support decisions in closed-loop driving.

\bibliographystyle{IEEEtran}
\bibliography{ref}

\end{document}